\documentclass[10pt,twocolumn,letterpaper]{article}
\usepackage[utf8]{inputenc} 
\usepackage{cvpr}
\usepackage{times}
\usepackage{epsfig}
\usepackage{graphicx}
\usepackage{amsmath}
\usepackage{amssymb}
\usepackage[mode=buildnew]{standalone}
\usepackage[font=small]{subfig}
\usepackage{booktabs}
\usepackage{multirow}
\usepackage{siunitx}
\usepackage[english]{babel}
\usepackage{xcolor}

\usepackage[pagebackref=true,breaklinks=true,letterpaper=true,colorlinks,bookmarks=false]{hyperref}

\cvprfinalcopy 

\def\cvprPaperID{11} 

\ifcvprfinal\fi
\begin{document}

\title{Self-Supervised Domain Mismatch Estimation for Autonomous Perception}

\author{Jonas Löhdefink$^1$ \quad Justin Fehrling$^1$ \quad Marvin Klingner$^1$ \quad Fabian Hüger$^2$ \quad Peter Schlicht$^2$ \\[0.5em]
	Nico M. Schmidt$^2$  \quad Tim Fingscheidt$^1$\\[0.5em]
	{\tt\small \{j.loehdefink, j.fehrling, m.klingner, t.fingscheidt\}@tu-bs.de}\\
	{\tt\small \{fabian.hueger, peter.schlicht, nico.maurice.schmidt\}@volkswagen.de}\\
	\and
	$^1$Technische Universität Braunschweig \hspace{2cm} $^2$Volkswagen Group Automation
}

\maketitle

\begin{abstract}
Autonomous driving requires self awareness of its perception functions.
Technically spoken, this can be realized by observers, which monitor the performance indicators of various perception modules.
In this work we choose, exemplarily, a semantic segmentation to be monitored, and propose an autoencoder, trained in a self-supervised fashion on the very same training data as the semantic segmentation to be monitored.
While the autoencoder's image reconstruction performance (PSNR) during online inference shows already a good predictive power w.r.t. semantic segmentation performance, we propose a novel domain mismatch metric DM as the earth mover's distance between a pre-stored PSNR distribution on training (source) data, and an online-acquired PSNR distribution on any inference (target) data.
We are able to show by experiments that the DM metric has a strong rank order correlation with the semantic segmentation within its functional scope.
We also propose a training domain-dependent threshold for the DM metric to define this functional scope.
\end{abstract}
\vspace{-0.5cm}

\section{Introduction}
\label{sec:intro}
Semantic segmentation is an essential function concerning camera-based perception for autonomous driving.
Because of its highly safety-critical nature, it is crucial to observe the performance during inference.
Domain shifts in the input space of images are one of the various issues that come into play, being part of everyday scenarios and must be handled.
These domain shifts could be, \eg, changing lighting or weather conditions such as rain or fog.
The first step towards a better assessment of the input domain is to detect and measure an occurring domain shift.

\begin{figure}[t!]
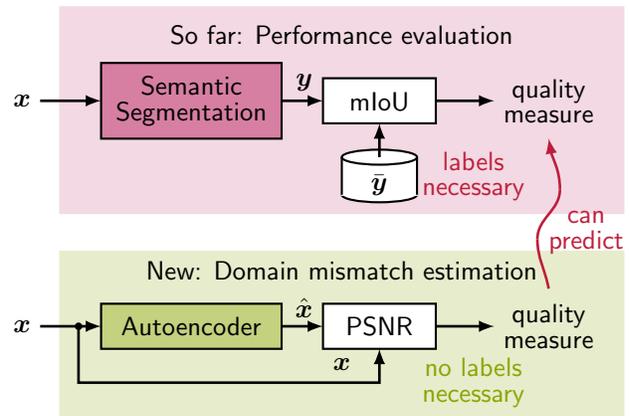

	\centering
	\includestandalone{figures/inference}
	\caption{
		\textbf{Performance evaluation of semantic segmentation} (simplified sketch).
		Evaluation of the mean intersection over union (mIoU) requires ground truth segmentation labels $\bar{\boldsymbol{y}}$, while the proposed domain mismatch estimation is performed on the basis of the PSNR of an autoencoder, trained and evaluated without labels.
	}
	\label{fig:inference}
	\vspace{-0.5cm}
\end{figure}

The commonly used quality measure for object detection and semantic segmentation is the mean intersection over union (mIoU).
Unfortunately, an mIoU can only be computed with ground truth semantic segmentation labels at hand, which are not available online during driving, of course.
Besides semantic segmentation networks, we assume that other learned functions (even for different tasks) also perform worse when it comes to a performance degradation of the segmentation caused by a domain shift, assuming they were trained on the same data distribution.
Hence, we propose the use of a (self-supervised) autoencoder, which allows to monitor domain shifts by computation of a peak-signal-to-noise ratio (PSNR) between input and output images without requiring labels, see Figure~\ref{fig:inference}.
Clearly, it is difficult to determine the domain shift on single images as there may always be unusual images, so we focus on investigating batches of images.
In fact, we train and evaluate the framework on various datasets simulating domain shifts.
A first simple approach to estimate the domain shift is to evaluate the resulting autoencoder's mean PSNR scores.
We also compute PSNR performance histograms both for the training data and for different inference data domains and compare them by the earth mover's distance (EMD)~\cite{Rubner2000}, obtaining a domain mismatch (DM) metric between two datasets.
In our experimental evaluation, we evaluate the PSNR and our novel DM metric with the absolute segmentation performance difference in mIoU, showing a strong correlation for both.

The rest of this paper is structured as follows:
Section~\ref{sec:rel_work} presents an overview of the state of the art for related fields of research.
In Section~\ref{sec:domain_mismatch_estimation}, we explain the details of our domain mismatch estimation.
Section~\ref{sec:discussion} then discusses and interprets the results of the conducted experiments.
Finally, we conclude our findings in Section~\ref{sec:conclusion}.

\section{Related Work}
\label{sec:rel_work}
In this section we provide an overview of the most re\-le\-vant state-of-the-art approaches of semantic segmentation, autoencoders, and domain shift.

\textbf{Semantic Segmentation}
can be considered as pixel-wise classification of images.
Some areas of applications for semantic segmentation are medical image analysis, perception of autonomous driving~\cite{Baer2019, Bolte2019}, video surveillance, and augmented reality~\cite{minaee2020image}.

The architectural concepts for semantic segmentation can be categorized into fully convolutional networks (FCNs)~\cite{Long2015}, graphical models~\cite{Chen2015}, encoder-decoder based models~\cite{Noh2015}, multi-scale architectures~\cite{lin2017feature}, region CNNs (R-CNNs)~\cite{He2017}, networks based on dilated convolutions~\cite{Chen2015, Chen2017}, recurrent neural networks (RNNs)~\cite{visin2016reseg}, attention-based models~\cite{Chen2016a}, generative adversarial networks (GANs)~\cite{Goodfellow2014, Luc2016}, and active contour models~\cite{kass1988snakes}, as comprehensively investigated in~\cite{minaee2020image}.
Furthermore, there is also a variety of image segmentation datasets in 2D, 2.5D (including depth), and 3D.
Often used 2D datasets are PASCAL VOC \cite{Everingham2015}, PASCAL VOC12 \cite{everingham2010pascal}, MS COCO \cite{lin2014microsoft}, Cityscapes \cite{Cordts2016}, KITTI \cite{Geiger2013}, SYNTHIA \cite{ros2016synthia}, Berkeley DeepDrive~\cite{Yu2018b}, and CamVid~\cite{brostow2009}.
For the evaluation of semantic segmentation models, se\-ve\-ral quality measures are frequently used, e.g., pixel accuracy (PA), mean pixel accuracy (MPA), mean intersection over union (mIoU), precision, F1-score, and dice coefficient~\cite{minaee2020image}.

Due to the efficient implementation and therefore also training and inference time savings, we use the encoder-decoder-based ERFNet~\cite{Romera2018}, which adopts its architecture from~\cite{Paszke2016} and \cite{Badrinarayanan2016}.
For our experiments, we use the Cityscapes dataset~\cite{Cordts2016}, the KITTI dataset~\cite{Geiger2013} and the Berkeley DeepDrive dataset~\cite{Yu2018b} and report the mIoU since it is the most wide-spread segmentation metric.

\begin{figure*}[t!]
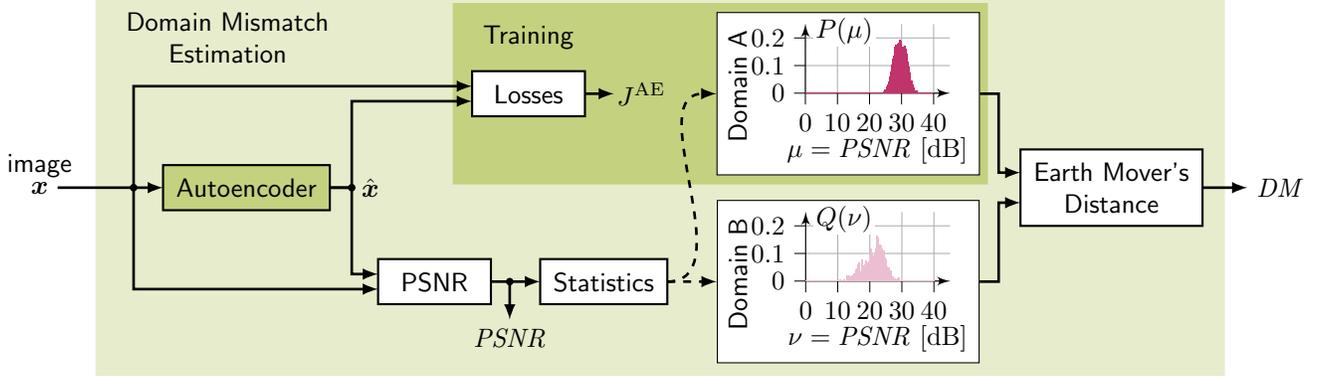

	\centering
	\includestandalone{figures/domain_mismatch_estimation}
	\caption{\textbf{Our proposed domain mismatch estimation}.
		The loss function for the autoencoder is only used during self-supervised training and is not needed during inference.
		The histogram of $\mathit{PSNR}$ values in the training data domain (A) is compared to an acquired histogram during inference (domain B), using the earth mover's distance (EMD), yielding the proposed domain mismatch metric $\mathit{DM}$.}
	\label{fig:domain_mismatch_estimation}
\end{figure*}

\textbf{Autoencoders}
are a special case of encoder-decoder architectures, trained to have the same input and output in a self-supervised fashion.
Variations of autoencoders can be found in their respective architectures, loss functions, learning principles, and strategies.

Due to the bottleneck in the autoencoder, it is inherently closely related to image compression~\cite{Agustsson2018, Loehdefink2019, Theis2017}, which often adds quantization, and also to image (and video) super-resolution (SR) methods~\cite{haris2018deep, mao2016image}, focussing on reconstructing the original high-resolution image from a low-resolution representation.
Furthermore, also texture synthesis~\cite{li2016precomputed, ulyanov2017improved}, image inpainting~\cite{yeh2017semantic, yu2018generative}, and style transfer~\cite{gatys2016image, Karras2019} incorporate autoencoder structures.
In many cases, decoders make use of transposed convolutions~\cite{zeiler2014visualizing, zeiler2010deconvolutional} and multi-task learning~\cite{caruana1997multitask, He2017}.
Besides this, many architectures use generative adversarial networks (GANs)~\cite{Goodfellow2014} or ex\-ten\-sions such as the conditional GAN (cGAN)~\cite{Mirza2014}, or the least squares GAN (LSGAN)~\cite{Mao2016}.
The Wasserstein GAN (WGAN)~\cite{Arjovsky2017} is another famous representative of GANs, using the Wasserstein-1 distance, also known as the earth mover's distance (EMD)~\cite{Rubner2000}, which we will use as domain mismatch metric.
Commonly used quality measures for image compression systems, super resolution approaches, and autoencoders in general are peak-signal-to-noise ratio (PSNR)~\cite{Loehdefink2019, Salomon2004}, structural si\-mi\-la\-ri\-ty (SSIM)~\cite{wang2004image}, and multi-scale SSIM (MS-SSIM)~\cite{wang2003multi}, as well as the mean opinion score (MOS), which is the human-evaluated perceptual quality.
Besides, there are numerous other image quality assessment methods, trying to simulate the human perception system~\cite{ma2017learning, talebi2018nima}.

We use the autoencoder architecture for learned image compression from~\cite{Agustsson2018}, with the difference that we omit the quantization block, since we do not aim at compression.

\textbf{Domain Shift}
deals with variations between data domains or distributions, while domains can be considered as environments of different technical or natural data characteristics and different data distributions.
Examples for such domain shifts are differing sensor setups in capture devices, or traffic signs in different countries.

Learning models on data distributions differing from the application distributions is referred to as transfer learning~\cite{Pan2010, Venkateswara2017}, since the goal is to transfer the learned know\-ledge.
Specifically, domain adaptation approaches~\cite{Bolte2019a, Ganin2015} aim at adjusting models to perform well in two (or more) domains in a (semi-)supervised or unsupervised fashion.
Moreover, time-variant domains often lead to conceptual drifts~\cite{tsymbal2004problem, widmer1996learning}, posing a particularly difficult problem, since the direction of the drift is unknown.
This makes the drift even more important to detect.
The maximum mean discrepancy (MMD)~\cite{borgwardt2006integrating, gretton2012kernel} is another task-independent method to measure a domain shift between a source and a target domain.
In this technique, a function in a reproducing kernel Hilbert space (RKHS) is to be found, being large for samples from the first distribution $p$ and small for samples from the second distribution $q$.
The MMD then is computed by subtracting the mean of function outputs with inputs from $q$ from the mean of function outputs with inputs from $p$.
This method can be thought of comparing not only the means of two distributions but also their higher order moments such as the variance.

The main differences between the MMD and our method is that, first, the MMD maximizes the sample expectation differences from two distributions in a reproducing kernel Hilbert space over a set of functions for each domain pair to be evaluated, while our proposed method is trained only once on the training (source) domain.
Second, the MMD uses the difference of mean values to obtain the final metric, while we evaluate the outputs by the EMD.
And third, we use neural networks both for semantic segmentation and for domain mismatch estimation, while the function optimized in the typical MMD is not related to neural networks.

\section{Domain Mismatch Estimation}
\label{sec:domain_mismatch_estimation}
A detailed block diagram of the proposed domain mismatch estimation can be seen in Figure~\ref{fig:domain_mismatch_estimation}.
It consists of an autoencoder along with a loss function and computational steps to obtain a domain mismatch metric DM.
The image $\boldsymbol{x} = (\boldsymbol{x}_i)$ with height $H$ and width $W$, consisting of normalized (color) pixels $\boldsymbol{x}_i\in [-1, 1]^C$, with $C=3$ color channels and pixel index \mbox{$i\in \mathcal{I} = \{1,2,..., H\!\cdot\!W\}$}, is the input to both, an undisplayed but to be observed semantic segmentation, and to our proposed domain mismatch estimator.
Its autoencoder receives the normalized image $\boldsymbol{x}$ and produces an image reconstruction $\hat{\boldsymbol{x}} = (\hat{\boldsymbol{x}}_i)$ with $\hat{\boldsymbol{x}}_i \in [-1, 1]^C$.
An advantage of all autoencoder settings is the fact that no explicit labels are needed because of its self-supervised training.
So in addition to the image reconstruction, the loss and quality measure also use the input image $\boldsymbol{x}$.
Different domains result in different self-supervised quality measure distributions, which can then be compared by the earth mover's distance~\cite{Rubner2000}, providing our proposed domain mismatch metric.

\subsection{Network Architectures and Losses}
We use the ERFNet~\cite{Romera2018} for the task of semantic segmentation to be observed.
The network is optimized to run in real-time, while still achieving accurate results.
It has an encoder/decoder structure and makes use of factorized residual layers consisting of a combination of two 1D filters instead one 2D filter.
Since the semantic segmentation architecture and loss function are identical to that used in \cite{Romera2018}, we refer the interested reader to this reference.

Concerning our autoencoder, we use an adversarial archi\-tec\-ture adopted from~\cite{Agustsson2018}, \cite{wang2018high}, and \cite{Johnson2016}.
Speaking in terms of a generative adversarial network, the generator combines the encoder and decoder networks of the autoencoder and the discriminator evaluates its reconstructions in a simultaneous training.
In the encoder, decoder, and discriminator, each convolutional operation is zero-padded, always preserving the image dimensions, and followed by an instance normalization layer as well as a ReLU activation function if not stated otherwise.

First in the encoder, there is a convolutional layer with kernel size $7\times 7$, stride of $1$, and 60 feature maps.
After\-wards, 4 downsampling blocks follow, each consisting of a convolutional layer with kernel size $3\times 3$, and a stride of two for spatial reduction of the $(120, 240, 480, 960)$ feature maps.
The last convolutional layer has a kernel size $3\times 3$, stride of one, and 8 feature maps, shaping the bottleneck.
The final encoder layer has a tanh activation to yield outputs in the range $[-1, 1]$.

The decoder architecture first has a convolutional layer with kernel size $3\times 3$, stride of one, and 960 feature maps.
Afterwards, there are 9 residual blocks, each consisting of two convolutional layers, bypassed by an identity function, where the second convolutional layer omits the ReLU activation function.
The initial image resolution is restored by 4 transposed convolutional layers with kernel size $4\times 4$, stride of two, and $(960, 480, 240, 120)$ feature maps.
The architecture is finalized by a convolutional layer with kernel size $7 \times 7$, stride of $1$, three feature maps, and a tanh activation function.

In the discriminator, instead of the ReLU activation function, the leakyReLU function is used.
The discriminator consists of 4 convolutional layers with kernel size $4\times 4$, stride of $2$, and $(64, 128, 256, 512)$ feature maps.
A final convolutional layer with kernel size $4 \times 4$, stride of one, one feature map, and ReLU activation delivers the discriminator outputs.

The autoencoder loss
\begin{equation}
\label{eq:loss}
J^\mathrm{AE} = \alpha_1 J^\mathrm{dist} + \alpha_2 J^\mathrm{FM} + (1-\alpha_1-\alpha_2) J^\mathrm{G, adv},
\end{equation}
with the weighting factors \mbox{$\alpha_1,\alpha_2\!\in\! [0,1], \alpha_1 + \alpha_2 \le 1$}, consists of an MSE distortion loss $J^\mathrm{dist}$, the L1 feature map loss $J^\mathrm{FM}$ between the discriminator's feature activations fed with the image $\boldsymbol{x}$ and the reconstruction $\hat{\boldsymbol{x}}$, and the generator-specific least-squares (LS) GAN loss $J^\mathrm{G, adv}$~\cite{Mao2016}.
The discriminator is trained with the discriminator-specific LS-GAN loss $J^\mathrm{D, adv}$, which pursues the opposed goal of the generator.

\subsection{Quality Measures}
Evaluating the semantic segmentation performance for a set of images, commonly the mean intersection over union
\begin{equation}
\label{eq:miou}
\mathit{mIoU} = \frac{1}{|\mathcal{S}|} \sum_{s\in \mathcal{S}}\frac{\mathit{TP_s}}{\mathit{TP_s} + \mathit{FP_s} + \mathit{FN_s}}
\end{equation}
is used, being composed of the numbers of true-positive ($\mathit{TP_s}$) pixels, false-positive ($\mathit{FP_s}$) pixels, and false-negative ($\mathit{FN_s}$) pixels w.r.t. the ground truth, with the class index $s\in \mathcal{S} = \{1, 2, ..., S\}$, being summed up over all images before.

\begin{table*}[t]
	\centering
	\begin{tabular}{ccccccccc}
		\toprule
		     \multirow{2}{*}[-2pt]{Trained on}      & \multirow{2}{*}[-2pt]{Model} & \multirow{2}{*}[-2pt]{Measure} &                                         \multicolumn{5}{c}{Evaluated on}                                         & \multirow{2}{*}[-2pt]{Kendall $\tau$} \\
		          \cmidrule(l{2pt}){4-8}            &                              &                                & CS$_\textrm{train}$  &  CS$_\textrm{val}$   & BDD$_\textrm{train}$ &  BDD$_\textrm{val}$  &        KITTI         &                                       \\ \midrule
		\multirow{2}{*}[-2pt]{CS$_\textrm{train}$}  &         Autoencoder          &              PSNR              & \SI{29.55}{\decibel} & \SI{28.24}{\decibel} & \SI{21.01}{\decibel} & \SI{21.26}{\decibel} & \SI{20.13}{\decibel} &      \multirow{2}{*}[-2pt]{0.6}       \\
		          \cmidrule(l{2pt}){4-8}            &         Segmentation         &              mIoU              & \SI{81.2}{\percent}  & \SI{66.7}{\percent}  & \SI{23.1}{\percent}  & \SI{26.7}{\percent}  & \SI{51.1}{\percent}  &                                       \\ \midrule
		\multirow{2}{*}[-2pt]{BDD$_\textrm{train}$} &         Autoencoder          &              PSNR              & \SI{25.18}{\decibel} & \SI{25.13}{\decibel} & \SI{25.87}{\decibel} & \SI{25.37}{\decibel} & \SI{22.10}{\decibel} &      \multirow{2}{*}[-2pt]{0.8}       \\
		          \cmidrule(l{2pt}){4-8}            &         Segmentation         &              mIoU              & \SI{45.5}{\percent}  & \SI{43.9}{\percent}  & \SI{53.8}{\percent}  & \SI{49.0}{\percent}  & \SI{44.1}{\percent}  &                                       \\ \bottomrule
	\end{tabular}
	\caption{Mean PSNR results for the autoencoder and mIoU results for the semantic segmentation trained and evaluated on various datasets.}
	\label{tab:domain_performances}
\end{table*}

For the evaluation of the autoencoder, the image reconstruction quality for input and output color image pixels in the number range $\boldsymbol{x}^\prime_i, \hat{\boldsymbol{x}}^\prime_i \in [0,255]^C$ usually is computed by the peak signal-to-noise ratio (PSNR), performing a direct MSE comparison of pixel values:
\begin{equation}
\label{eq:psnr}
\mathit{PSNR} = 10 \log\left(\frac{(x^\prime_\textrm{max})^2}{\frac{1}{C \cdot H \cdot W}\sum_{i\in\mathcal{I}} \left\lVert \boldsymbol{x}^\prime_i-\hat{\boldsymbol{x}}^\prime_i\right\rVert^2}\right) [\si{\decibel}]
\end{equation}
with ${x^\prime_\textrm{max}} = 255$.

The comparison of two discrete probability distributions $P(\mu)$, \mbox{$\mu \in \mathcal{M} = \{1,2,...,M\}$} and $Q(\nu)$, \mbox{$\nu \in \mathcal{N} = \{1,2,...,N\}$} can be computed by the earth-mover's distance (EMD)~\cite{Rubner2000}.
This metric computes the minimum work $W$ required to convert one distribution into the other by multiplying the distance $d_{\mu \nu} = |\mu - \nu| \in \{0,1,...,\max(M,N)\!-\!1\}$ bet\-ween the bins with index $\mu$ and $\nu$ with the $M\times N$ flow matrix $\boldsymbol{F} = (f_{\mu \nu})$ with $f_{\mu \nu}\in [0,1]$ being the \textit{flow} from bin $\mu$ to $\nu$.
The optimal flow is found by minimizing the work according to
\begin{equation}
\boldsymbol{F}^*=\arg\min\limits_{\boldsymbol{F}} W(P, Q, \boldsymbol{F}) = \arg\min\limits_{\boldsymbol{F}} \sum_{\mu\in\mathcal{M}}\sum_{\nu\in\mathcal{N}}f_{\mu\nu}d_{\mu\nu}
\end{equation}
under consideration of the four (stochastic) constraints
\begin{align*}
	f_{\mu\nu}                                                      & \geq 0,                                         & \mu\in\mathcal{M}, \nu\in\mathcal{N} \\
	\sum_{\nu\in \mathcal{N}}f_{\mu\nu}                             & \leq P(\mu),                                    & \mu\in\mathcal{M}                    \\
	\sum_{\mu\in \mathcal{M}}f_{\mu\nu}                             & \leq Q(\nu),                                    & \nu\in\mathcal{N}                    \\
	\sum_{\mu\in \mathcal{M}} \sum_{\nu \in \mathcal{N}} f_{\mu\nu} & = \operatorname{min}(P(\mu), Q(\nu)).
\end{align*}
We then obtain the earth-mover's distance as
\begin{equation}
\label{eq:emd}
\mathit{DM}(P, Q) = \frac{\sum\limits_{\mu \in \mathcal{M}} \sum\limits_{\nu \in \mathcal{N}} f^*_{\mu \nu} d_{\mu \nu}}{\sum\limits_{\mu \in \mathcal{M}} \sum\limits_{\nu \in \mathcal{N}} f^*_{\mu \nu}},
\end{equation}
which we will use as our proposed domain mismatch metric by computing the difference of reconstruction qualities for various datasets.

We use Kendall's rank order coefficient \cite{Abel2016, kendall1945treatment} \mbox{$\tau = \tau_\textrm{b}$}, which accounts for ties in one quantity, whereby in the following we will omit the index b.
Having $K$ observations \mbox{$\boldsymbol{o}_k = (a_k, b_k)$} with $k\in \{1,..., K\}$, the total number of observation pairs
\begin{equation}
(\boldsymbol{o}_k, \boldsymbol{o}_\ell) = \bigl((a_k, b_k), (a_\ell, b_\ell)\bigr)
\end{equation}
with $k<\ell$ is \mbox{$n_\textrm{p} = \binom{K}{2} = \frac{1}{2}K(K\!-\!1)$}.
A pair of observations is called \textit{concordant} if the observation's components have the same order (both ascending or both descending), otherwise it is \textit{discordant}.
If the values of one component in the pair are equal, it is called a tie in this component (here: a tie in $a$ or a tie in $b$) and is neither concordant nor discordant.
The number of concordant pairs $n_\textrm{c}$, discordant pairs $n_\textrm{d}$, ties in $a$ $n_\textrm{a}$, and ties in $b$ $n_\textrm{b}$ is used to calculate Kendall's rank order coefficient
\begin{equation}
\label{eq:tau}
\tau = \frac{n_\textrm{c} - n_\textrm{d}}{\sqrt{(n - n_\textrm{a})(n - n_\textrm{b})}} \in [-1, 1],
\end{equation}
where $\tau = 1$ means that the observations are perfectly in the same order, $\tau = -1$ means that they are perfectly in reversed order, and $\tau = 0$ means that there is no correlation in rank order.

\section{Evaluation and Discussion}
\label{sec:discussion}
In this section, we will introduce the training setup and describe the performance of the segmentation and autoencoder networks on different datasets, as well as we will analyze the proposed method for domain mismatch estimation.

\subsection{Data Configurations and Training}
For experimental evaluation, we use Cityscapes~\cite{Cordts2016}, containing images from several German cities, Berkeley DeepDrive~\cite{Yu2018b}, containing data from the U.S., and KITTI~\cite{Geiger2013}, containing data from a single German city including surroundings.
All these datasets provide the same class labeling scheme for segmentation and are therefore compatible.
Furthermore, they all provide a training and a validation set with segmentation labels.
For our experiments we distinguish between the Cityscapes training set (CS$_\textrm{train}$), the Cityscapes validation set (CS$_\textrm{val}$), the Berkeley DeepDrive training set (BDD$_\textrm{train}$), the Berkeley DeepDrive validation set (BDD$_\textrm{val}$), and the KITTI set (which consists of all first images in the stereo training set of KITTI2015).
CS$_\textrm{train}$ and CS$_\textrm{val}$ consists of 2,975 and 500 images, respectively, and are downsampled to $512 \times 1024$ pixels.
BDD$_\textrm{train}$ and BDD$_\textrm{val}$ have 7,000 and 1,000 images, respectively, with a resolution of $1280 \times 720$ pixels.
Finally, the KITTI training split has 200 images with a resolution of $375 \times 1242$ pixels.
The models for the semantic segmentation and the autoencoder are trained with \texttt{PyTorch}~\cite{Paszke2017} either with CS$_\textrm{train}$ or BDD$_\textrm{train}$ on an \texttt{NVidia GTX 1080 Ti} GPU.

The encoder of the segmentation network is pretrained on Image\-Net~\cite{Russakovsky2015}.
For data augmentation, the training images are randomly flipped horizontally and cropped to $192 \times 640$ pixels.
After the pretraining, we continue training for 200 epochs with a batch size of 6, an initial learning rate of 0.0005, an Adam optimizer~\cite{Kingma2015} with $\beta_1=0.9$ and $\beta_2=0.999$, and a weight decay of 0.0002, while ignoring the background class.

The GAN training procedure first optimizes the generator while fixing the discriminator weights, and vice versa afterwards.
We train for $50$ epochs with batch size $1$, and an initial learning rate of $0.0002$, using an Adam optimizer with $\beta_1=0.5$ and $\beta_2=0.999$.
Concerning the autoencoder loss function~\eqref{eq:loss}, we use the weighting factors $\alpha_1=\frac{12}{23}$ for the MSE loss and $\alpha_2=\frac{10}{23}$ for the feature matching loss.
Furthermore, early stopping w.r.t.\ the PSNR on the validation set is applied.

\subsection{Domain-Specific Performance}
In this section, we first evaluate the performance of semantic segmentation and autoencoder individually with mIoU~\eqref{eq:miou} and PSNR~\eqref{eq:psnr}, respectively, for the different datasets.
The results for the models trained on CS$_\textrm{train}$ and BDD$_\textrm{train}$ can be seen in Table~\ref{tab:domain_performances}.
We also report Kendall's rank order coefficient $\tau$~\eqref{eq:tau}, evaluating the degree of rank si\-mi\-la\-ri\-ty of the PSNR and mIoU series.

For the CS$_\textrm{train}$-trained autoencoder, the PSNR performance is best on CS$_\textrm{train}$ (obviously because it is the training set) and performs second best on CS$_\textrm{val}$, which is also plausible since it is the in-domain case.
Evaluated on BDD$_\textrm{train}$ and BDD$_\textrm{val}$ the PSNR falls by several dB compared to the source domain to $\SI{21.01}{\decibel}$ and $\SI{21.26}{\decibel}$, respectively, due to the domain shift.
The lowest performance is achieved on KITTI with $\SI{20.13}{\decibel}$.
We observe a si\-mi\-lar ranking of performances in the semantic segmentation results of the segmentation trained on CS$_\textrm{train}$, with the surprising exception that the KITTI dataset this time does not yield the largest drop in mIoU.
When comparing rank orders, only the positions of BDD$_\textrm{val}$ and KITTI seem to be swapped.
The rank order coefficient $\tau \in [-1,1]$ is $0.6$, still indicating a po\-si\-tive correlation in the behavior of PSNR and mIoU.
Conclusively, we observe a huge domain-shift-induced performance drop for both models trained on the Cityscapes data, and evaluated on BDD and KITTI data.

As before, the autoencoder trained on BDD$_\textrm{train}$ performs best in its own domain with a PSNR of \SI{25.87}{\decibel} on the training set and \SI{25.37}{\decibel} on the validation set.
Evaluation on CS$_\textrm{train}$ and CS$_\textrm{val}$ is ranked third and fourth w.r.t.\ PSNR, even though the dB difference to the source domain is quite small.
The performance on KITTI is again lower than on the other datasets.
In the semantic segmentation, the mIoU again is best for the in-domain datasets BDD$_\textrm{train}$ and BDD$_\textrm{val}$, while CS$_\textrm{train}$, CS$_\textrm{val}$, and KITTI achieve si\-mi\-lar mIoU, which is a bit in contrast to the autoencoder performance, which indicates that KITTI has a larger domain shift than the others.
Kendall's $\tau$ is $0.8$, underlining the strong correlation of rank orders.

\begin{figure*}[t!]
	\centering
	\subfloat[Evaluated on CS$_\textrm{train}$,\newline trained on CS$_\textrm{train}$\label{fig:histogram_cstrain_cstrain}]{
		\includestandalone[]{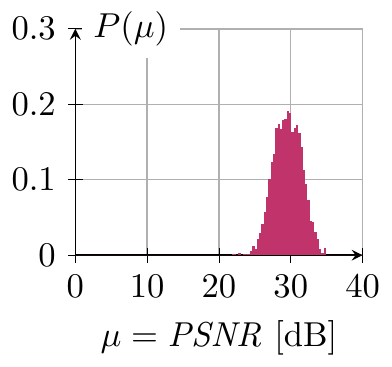}
	}
	\subfloat[Evaluated on CS$_\textrm{val}$,\newline trained on CS$_\textrm{train}$\label{fig:histogram_cstrain_csval}]{
		\includestandalone[]{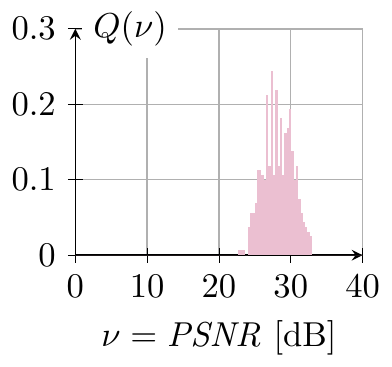}
	}
	\subfloat[Evaluated on BDD$_\textrm{val}$,\newline trained on CS$_\textrm{train}$\label{fig:histogram_cstrain_bddval}]{
		\includestandalone[]{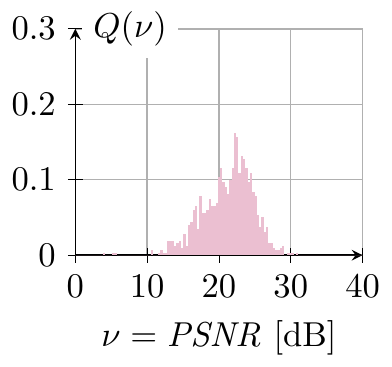}
	}
	\subfloat[Evaluated on KITTI,\newline trained on CS$_\textrm{train}$\label{fig:histogram_cstrain_kitti}]{
		\includestandalone[]{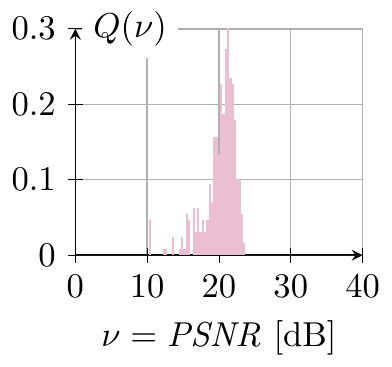}
	}
	\\
	\centering
	\subfloat[Evaluated on BDD$_\textrm{train}$,\newline trained on BDD$_\textrm{train}$\label{fig:histogram_bddtrain_bddtrain}]{
		\includestandalone[]{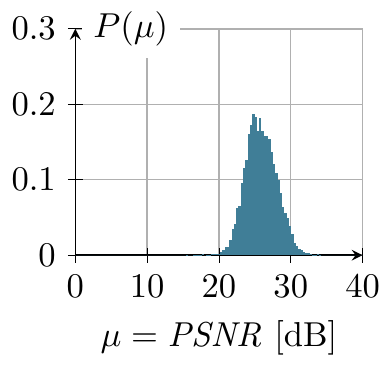}
	}
	\subfloat[Evaluated on BDD$_\textrm{val}$,\newline trained on BDD$_\textrm{train}$\label{fig:histogram_bddtrain_bddval}]{
		\includestandalone[]{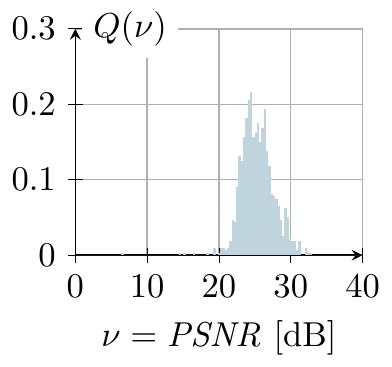}
	}
	\subfloat[Evaluated on CS$_\textrm{val}$,\newline trained on BDD$_\textrm{train}$\label{fig:histogram_bddtrain_csval}]{
		\includestandalone[]{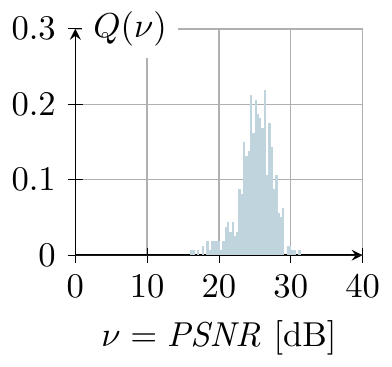}
	}
	\subfloat[Evaluated on KITTI,\newline trained on BDD$_\textrm{train}$\label{fig:histogram_bddtrain_kitti}]{
		\includestandalone[]{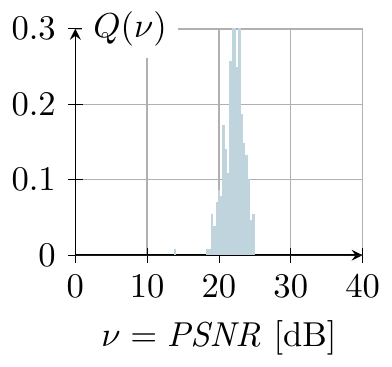}
	}
	\caption{Histograms $P(\mu)$ (source domain) and $Q(\nu)$ (target domain), with $\mu$, $\nu$, representing autoencoder performance PSNRs with models trained and evaluated on different datasets.
		The upper histograms (red, \ref{fig:histogram_cstrain_cstrain} to \ref{fig:histogram_cstrain_kitti}) stem from the autoencoder trained on CS$_\textrm{train}$, while the lower ones (blue, \ref{fig:histogram_bddtrain_bddtrain} to \ref{fig:histogram_bddtrain_kitti}) are trained on BDD$_\textrm{train}$.}
	\label{fig:histograms}
	\vspace*{-0.3cm}
\end{figure*}

The models trained on CS$_\textrm{train}$ and BDD$_\textrm{train}$ show at least si\-mi\-lar trends in both of the investigated tasks (autoencoder and segmentation), which encourages us to assign the autoencoder the role of an observer for the semantic segmentation.
The general trend is: Once PSNR drops, also mIoU can be assumed to drop, while the achievable absolute PSNR scores are data-dependent.
This makes it a bit tedious to define a threshold for an acceptable domain shift, since it varies for each training dataset.
Rank orders are not necessarily kept in the low PSNR regime (BDD$_\textrm{train}$, BDD$_\textrm{val}$, KITTI) for models trained on CS$_\textrm{train}$, and (CS$_\textrm{val}$, KITTI) for models trained on BDD$_\textrm{train}$:
However, even here we can reliably always assume that mIoU drops as well to unacceptable low values.
Already in this preliminary experiment, investigating mean performance scores, we observed that \textit{if semantic segmentation performance (mIoU) drops below training or validation set performance, also autoencoder performance (PSNR) drops.}

\subsection{Domain Mismatch}
For better visualization of domains, Figure~\ref{fig:histograms} shows PSNR histograms, resulting from the evaluation on the individual datasets.
For both source domains CS and BDD, evaluating the training set itself yields smooth distributions of PSNR scores around their mean values as expected (almost Gaussian), see Figures~\ref{fig:histogram_cstrain_cstrain} and \ref{fig:histogram_bddtrain_bddtrain}.
The transition to the validation set in the source domain and further on to one of the target domains implies a decrease of the mean PSNR and an increase of the standard deviation in the distribution, as can be seen in the Figures~\ref{fig:histogram_cstrain_csval} to \ref{fig:histogram_cstrain_kitti} for the CS-trained autoencoder, and in Figures~\ref{fig:histogram_bddtrain_bddval} to \ref{fig:histogram_bddtrain_kitti} for the BDD-trained autoencoder.
Noteworthy, the KITTI dataset is only from a single German city, which may be the cause for the small standard deviation in the histograms~\ref{fig:histogram_cstrain_kitti} and \ref{fig:histogram_bddtrain_kitti}.

\begin{table*}[t]
	\centering
	\begin{tabular}{cccccccccc}
		\toprule
		\multirow{2}{*}[-2pt]{Trained on}      &      \multirow{2}{*}[-2pt]{Reference}       & \multirow{2}{*}[-2pt]{Model} & \multirow{2}{*}[-2pt]{Measure} &                                       \multicolumn{5}{c}{Evaluated on}                                       & \multirow{2}{*}[-2pt]{Kendall $\tau$} \\
		\cmidrule(l{2pt}){5-9}            &                                             &                              &                                & CS$_\textrm{train}$ &  CS$_\textrm{val}$  & BDD$_\textrm{train}$ & BDD$_\textrm{val}$  &        KITTI        &                                       \\ \midrule
		\multirow{2}{*}[-2pt]{CS$_\textrm{train}$}  & \multirow{2}{*}[-2pt]{CS$_\textrm{train}$}  &         Autoencoder          &               DM               & \SI{0.0}{\decibel}  & \SI{1.31}{\decibel} & \SI{8.53}{\decibel}  & \SI{8.29}{\decibel} & \SI{9.41}{\decibel} &      \multirow{2}{*}[-2pt]{0.6}       \\
		\cmidrule(l{2pt}){5-9}            &                                             &         Segmentation         &          $\Delta$mIoU          & \SI{0.0}{\percent}  & \SI{14.5}{\percent} & \SI{58.1}{\percent}  & \SI{54.5}{\percent} & \SI{30.1}{\percent} &                                       \\ \midrule
		\multirow{2}{*}[-2pt]{BDD$_\textrm{train}$} & \multirow{2}{*}[-2pt]{BDD$_\textrm{train}$} &         Autoencoder          &               DM               & \SI{0.68}{\decibel} & \SI{0.74}{\decibel} &  \SI{0.0}{\decibel}  & \SI{0.51}{\decibel} & \SI{3.77}{\decibel} &      \multirow{2}{*}[-2pt]{0.8}       \\
		\cmidrule(l{2pt}){5-9}            &                                             &         Segmentation         &          $\Delta$mIoU          & \SI{8.3}{\percent}  & \SI{9.9}{\percent}  &  \SI{0.0}{\percent}  & \SI{4.8}{\percent}  & \SI{9.7}{\percent}  &                                       \\ \bottomrule
	\end{tabular}
	\caption{Domain mismatch metric $\mathit{DM}$~\eqref{eq:emd}, absolute mIoU differences between the references (CS$_\textrm{val}$ and BDD$_\textrm{val}$) and various datasets, and Kendall's rank order $\tau$.}
	\label{tab:domain_mismatches}
	\vspace*{-0.4cm}
\end{table*}

Table~\ref{tab:domain_mismatches} shows the mIoU differences and earth mover's distance (EMD) scores, namely our proposed domain mismatch scores DM~\eqref{eq:emd}, based on the PSNR histograms for the segmentation and the auto\-encoder, respectively.
Also, Kendall's rank order coefficient $\tau$ is provided, here evaluating the rank order si\-mi\-la\-ri\-ty of the DM and $\Delta$mIoU series.
The segmentation performance drop is simply stated as the mIoU difference between the training domains (CS$_\textrm{train}$ and BDD$_\textrm{train}$, respectively) and the target domains.

In consideration of the results for the Cityscapes-trained models, the DM metric for the validation set (here: $\SI{1.31}{\dB}$) indicates what is to be considered as default (or: typical) domain shift for in-domain data.
For each of the out-of-domain shifts, regardless whether the target domain is BDD$_\textrm{train}$, BDD$_\textrm{val}$, or KITTI, the autoencoder reconstruction performance dropped significantly, so our DM metric increased to \SI{8}{\decibel} and more.
In each of these cases also the drop in the mIoU is large, with $\Delta$mIoU being more than $\SI{50}{\percent}$ absolute for both BDD splits and $\SI{30.1}{\percent}$ for KITTI.
Again, the mIoU drop on KITTI is not the worst (although the DM metric is), but a \SI{30.1}{\percent} absolute mIoU drop definitely justifies KITTI to be ``out-of-domain'', as it is marked by the high $\mathit{DM} = \SI{9.41}{\decibel}$.
The pure rank orders in the DM metric and the $\Delta$mIoU series lead to a rank order coefficient $\tau$ of $0.6$, which is still indicating a positive rank correlation.
\ifcvprfinal
\else	
\pagebreak
\fi

Considering the models trained on BDD$_\textrm{train}$, the validation set domain shift of $\SI{0.51}{\decibel}$ is smaller than for Cityscapes, corresponding to an mIoU difference of $\SI{4.8}{\percent}$ to the training set.
The domain mismatch estimate DM for both CS datasets is a bit higher as with BDD$_\textrm{val}$, so we assume that DM and $\Delta$mIoU perform proportionally.
And indeed, as the DM metric increases from $\SI{0.51}{dB}$ for BDD$_\textrm{val}$ over $\SI{0.68}{dB}$ for CS$_\textrm{train}$ to $\SI{0.74}{dB}$ for CS$_\textrm{val}$, also the $\Delta$mIoU increases following the same rank order of datasets.
Interestingly again, the DM metric for KITTI is highest (here: by far highest), which is appropriate for $\Delta$mIoU being more than doubled w.r.t.\ the source validation set BDD$_\textrm{val}$.
Due to the concordant rank order of the DM metric and the $\Delta$mIoU in all but one cases, Kendall's rank order coefficient for the BDD-trained models is $0.8$.

We infer that \textit{the autoencoder is even more sensitive to domain shifts than the semantic segmentation}, since for both training datasets, the PSNR evaluated on KITTI dropped significantly while the mIoU showed a smaller decrease.
Nevertheless, for small values of our DM metric, the experiments show that the rank orders are concordant, as can especially be seen for the BDD-trained models.
Therefore, we propose to set a threshold for the DM metric to define its functional scope, in which the rank orders of the DM metric are expected to correspond to those of the $\Delta$mIoU.
The threshold should be \textit{two times the DM score of the in-domain validation set}, so it is depending on the specific domain it is trained and validated in.
Hence, for the CS-trained autoencoder the threshold lies at $2 \times \SI{1.31}{\decibel} = \SI{2.62}{\decibel}$, excluding BDD$_\textrm{train}$, BDD$_\textrm{val}$, and KITTI from the functional scope (meaning these are clearly out-of-domain datasets!), and for the BDD-trained autoencoder the treshold is $2 \times \SI{0.51}{\decibel} = \SI{1.02}{\decibel}$, which excludes only the KITTI dataset.
\textit{Inside its functional scope, the DM metric makes a statement about the semantic segmentation performance with concordant rank ordering.}
In comparison to the PSNR, \textit{we believe that the DM metric is the better generalizing metric, since the proposed threshold is relying on PSNR distributions, and is therefore less sensitive to single unusual images which do not yet necessarily make up a domain shift.}
As a result, the autoencoder is well-suited as a batch-type observer, since the DM metric exhibits reliable gradual estimations of the domain shift until exceeding the DM threshold, where the PSNR will collapse \textit{even before} the mIoU of the semantic segmentation.
DM results beyond the DM threshold always indicate a critical domain shift.

\section{Conclusions}
\label{sec:conclusion}
Observing the performance of safety-critical perception functions during autonomous driving is essential, because vehicles are by nature exposed to various environments, implying domain shifts.
We proposed a novel framework to monitor the quality of a semantic segmentation.
We accomplish this by estimating the domain shift by an autoencoder trained in self-supervised fashion.
A first approach is to evaluate mean PSNR scores which already show a strong rank order correlation to the mIoU.
However, comparing autoencoder outputs for various datasets by the earth mover's distance yields a more stable estimation of the domain shift which we propose as domain mismatch DM metric.
We found that the task of reconstructing an image is even more sensitive to domain shifts than semantic segmentation, being pixel-wise classification, which ultimately results in a certain functional scope for the autoencoder, beyond which input data can be clearly classified as ``out-of-domain''.
Within the valid functional scope of the autoencoder rank orders of our DM metric and mIoU differences are strongly rank-correlated.
The proposed DM metric is therefore shown to be well-suited as an observer.

\ifcvprfinal
\section*{Acknowledgment}
The research, leading to the results presented above, is funded by the German Federal Ministry for Economic Affairs and Energy within the project ``KI Absicherung – Safe AI for automated driving''.
\fi

{\small
	\bibliographystyle{ieee_fullname}
	\bibliography{ifn_spaml_bibliography}

\begin{thebibliography}{10}\itemsep=-1pt

\bibitem{Abel2016}
Johannes Abel, Magdalena Kaniewska, Cyril Guillaume, Wouter Tirry, and Tim
  Fingscheidt.
\newblock {Objective Assessment of Artificial Speech Bandwidth Extension
  Approaches}.
\newblock In {\em {Proc. of 12. ITG Symposium Speech Communication}}, pages
  190--194, Paderborn, Germany, Oct. 2016.

\bibitem{Agustsson2018}
Eirikur Agustsson, Michael Tschannen, Fabian Mentzer, Radu Timofte, and Luc~Van
  Gool.
\newblock {Generative Adversarial Networks for Extreme Learned Image
  Compression}.
\newblock In {\em Proc. of ICCV}, pages 221--231, Seoul, Korea, Oct. 2019.

\bibitem{Arjovsky2017}
Martin Arjovsky, Soumith Chintala, and Léon Bottou.
\newblock {Wasserstein GAN}.
\newblock In {\em Proc. of ICML}, pages 214--223, Sydney, Australia, Aug. 2017.

\bibitem{Badrinarayanan2016}
Vijay Badrinarayanan, Alex Kendall, and Roberto Cipolla.
\newblock {SegNet: A Deep Convolutional Encoder-Decoder Architecture for Image
  Segmentation}.
\newblock In {\em Proc. of PAMI}, pages 2481--2495, Kharagpur, India, Oct.
  2016.

\bibitem{Baer2019}
Andreas B\"{a}r, Fabian H\"{u}ger, Peter Schlicht, and Tim Fing\-scheidt.
\newblock {On the Robustness of Teacher-Student Frameworks for Semantic
  Segmentation}.
\newblock In {\em Proc. of CVPR - Workshops}, pages 1--9, Long Beach, CA, USA,
  June 2019.

\bibitem{Bolte2019}
Jan-Aike Bolte, Andreas B\"{a}r, Daniel Lipinski, and Tim Fing\-scheidt.
\newblock {Towards Corner Case Detection for Autonomous Driving}.
\newblock In {\em Proc. of IV}, pages 366--373, Paris, France, June 2019.

\bibitem{Bolte2019a}
Jan-Aike Bolte, Markus Kamp, Antonia Breuer, Silviu Homoceanu, Peter Schlicht,
  Fabian Huger, Daniel Lipinski, and Tim Fing\-scheidt.
\newblock {Unsupervised Domain Adaptation to Improve Image Segmentation Quality
  Both in the Source and Target Domain}.
\newblock In {\em Proc. of CVPR - Workshops}, pages 1--10, Long Beach, CA, USA,
  June 2019.

\bibitem{borgwardt2006integrating}
Karsten~M. Borgwardt, Arthur Gretton, Malte~J. Rasch, Hans-Peter Kriegel,
  Bernhard Sch{\"o}lkopf, and Alex~J. Smola.
\newblock {Integrating Structured Biological Data by Kernel Maximum Mean
  Discrepancy}.
\newblock {\em Bioinformatics}, 22(14):e49--e57, July 2006.

\bibitem{brostow2009}
Gabriel~J. Brostow, Julien Fauqueur, and Roberto Cipolla.
\newblock {Semantic Object Classes in Video: A High-Definition Ground Truth
  Database}.
\newblock {\em Pattern Recognition Letters}, 30(2):88--97, Jan. 2009.

\bibitem{caruana1997multitask}
Rich Caruana.
\newblock {Multitask Learning}.
\newblock {\em {Machine Learning}}, 28(1):41--75, July 1997.

\bibitem{Chen2015}
Liang-Chieh Chen, George Papandreou, Iasonas Kokkinos, Kevin Murphy, and
  Alan~L. Yuille.
\newblock {Semantic Image Segmentation With Deep Convolutional Nets and Fully
  Connected CRFs}.
\newblock In {\em Proc. of ICLR}, pages 1--14, San Diego, CA, USA, May 2015.

\bibitem{Chen2017}
Liang-Chieh Chen, George Papandreou, Florian Schroff, and Hartwig Adam.
\newblock {Rethinking Atrous Convolution for Semantic Image Segmentation}.
\newblock {\em arXiv}, June 2017.
\newblock (arXiv:1706.05587).

\bibitem{Chen2016a}
Liang-Chieh Chen, Yi Yang, Jiang Wang, Wei Xu, and Alan~L. Yuille.
\newblock {Attention to Scale: Scale-Aware Semantic Image Segmentation }.
\newblock In {\em Proc. of CVPR}, pages 1063--6919, Las Vegas, NV, USA, June
  2016.

\bibitem{Cordts2016}
Marius Cordts, Mohamed Omran, Sebastian Ramos, Timo Rehfeld, Markus Enzweiler,
  Rodrigo Benenson, Uwe Franke, Stefan Roth, and Bernt Schiele.
\newblock {The Cityscapes Dataset for Semantic Urban Scene Understanding}.
\newblock In {\em Proc. of CVPR}, pages 3213--3223, Las Vegas, NV, USA, June
  2016.

\bibitem{everingham2010pascal}
Mark Everingham, Luc Van~Gool, Christopher K.~I. Williams, John Winn, and
  Andrew Zisserman.
\newblock {The PASCAL Visual Object Classes (VOC) Challenge}.
\newblock {\em International Journal of Computer Vision}, 88(2):303--338, Sept.
  2010.

\bibitem{Everingham2015}
Mark Everingham, Luc Van~Gool, Christopher K.~I. Williams, John Winn, and
  Andrew Zisserman.
\newblock {The Pascal Visual Object Classes Challenge: A Retrospective}.
\newblock {\em International Journal of Computer Vision (IJCV)},
  111(1):98--136, Jan. 2015.

\bibitem{Ganin2015}
Yaroslav Ganin and Victor Lempitsky.
\newblock {Unsupervised Domain Adaptation by Backpropagation}.
\newblock In {\em Proc. of ICML}, pages 1180--1189, Lille, France, July 2015.

\bibitem{gatys2016image}
Leon~A. Gatys, Alexander~S. Ecker, and Matthias Bethge.
\newblock {Image Style Transfer Using Convolutional Neural Networks}.
\newblock In {\em Proc. of CVPR}, pages 2414--2423, Las Vegas, NV, USA, June
  2016.

\bibitem{Geiger2013}
Andreas Geiger, Philip Lenz, Christoph Stiller, and Raquel Urtasun.
\newblock {Vision Meets Robotics: The KITTI Dataset}.
\newblock {\em International Journal of Robotics Research (IJRR)},
  32(11):1231--1237, Aug. 2013.

\bibitem{Goodfellow2014}
Ian Goodfellow, Jean Pouget-Abadie, Mehdi Mirza, Bing Xu, David Warde-Farley,
  Sherjil Ozair, Aaron Courville, and Yoshua Bengio.
\newblock {Generative Adversarial Nets}.
\newblock In {\em Proc. of NIPS}, pages 2672--2680, Montr{\'{e}}al, Canada,
  Dec. 2014.

\bibitem{gretton2012kernel}
Arthur Gretton, Karsten~M. Borgwardt, Malte~J. Rasch, Bernhard Sch{\"o}lkopf,
  and Alexander Smola.
\newblock {A kernel two-sample test}.
\newblock {\em Journal of Machine Learning Research}, 13:723--773, Mar. 2012.

\bibitem{haris2018deep}
Muhammad Haris, Gregory Shakhnarovich, and Norimichi Ukita.
\newblock {Deep Back-Projection Networks for Super-Resolution}.
\newblock In {\em Proc. of CVPR}, pages 1664--1673, Salt Lake City, UT, USA,
  June 2018.

\bibitem{Johnson2016}
{Justin Johnson and Alexandre Alahi and Li Fei-Fei}.
\newblock {Perceptual Losses for Real-Time Style Transfer and
  Super-Resolution}.
\newblock In {\em Proc. of ECCV}, pages 694--711, Amsterdam, Netherlands, Oct.
  2016.

\bibitem{He2017}
{Kaiming He and Georgia Gkioxari and Piotr Dollár and Ross Girshick}.
\newblock {Mask R-CNN}.
\newblock In {\em Proc. of ICCV}, pages 2980--2988, Venice, Italy, Oct. 2017.

\bibitem{Karras2019}
Tero Karras, Samuli Laine, and Timo Aila.
\newblock {A Style-Based Generator Architecture for Generative Adversarial
  Networks}.
\newblock In {\em Proc. of CVPR}, pages 4401--4410, Long Beach, CA, USA, June
  2019.

\bibitem{kass1988snakes}
Michael Kass, Andrew Witkin, and Demetri Terzopoulos.
\newblock {Snakes: Active Contour Models}.
\newblock {\em International Journal of Computer Vision}, 1(4):321--331, Jan.
  1988.

\bibitem{kendall1945treatment}
Maurice~G. Kendall.
\newblock {The Treatment of Ties in Ranking Problems}.
\newblock {\em Biometrika}, 33(3):239--251, Nov. 1945.

\bibitem{Kingma2015}
Diederik~P. Kingma and Jimmy Ba.
\newblock {Adam: A Method for Stochastic Optimization}.
\newblock In {\em Proc. of ICLR}, pages 1--15, San Diego, CA, USA, May 2015.

\bibitem{li2016precomputed}
Chuan Li and Michael Wand.
\newblock {Precomputed Real-Time Texture Synthesis With Markovian Generative
  Adversarial Networks}.
\newblock In {\em Proc. of ECCV}, pages 702--716, Amsterdam, Netherlands, Oct.
  2016.

\bibitem{lin2017feature}
Tsung-Yi Lin, Piotr Doll{\'a}r, Ross Girshick, Kaiming He, Bharath Hariharan,
  and Serge Belongie.
\newblock {Feature Pyramid Networks for Object Detection}.
\newblock In {\em Proc. of CVPR}, pages 2117--2125, Honolulu, HI, USA, July
  2017.

\bibitem{lin2014microsoft}
Tsung-Yi Lin, Michael Maire, Serge Belongie, James Hays, Pietro Perona, Deva
  Ramanan, Piotr Doll{\'a}r, and C~Lawrence Zitnick.
\newblock {Microsoft COCO: Common Objects in Context}.
\newblock In {\em Proc. of ECCV}, pages 740--755, Zurich, Switzerland, Sept.
  2014.

\bibitem{Loehdefink2019}
Jonas {L{\"o}hdefink}, Andreas {B{\"a}r}, Nico~M. {Schmidt}, Fabian
  {H{\"u}ger}, Peter {Schlicht}, and Tim {Fing\-scheidt}.
\newblock {On Low-Bitrate Image Compression for Distributed Automotive
  Perception: Higher Peak SNR Does Not Mean Better Semantic Segmentation}.
\newblock In {\em Proc. of IV}, pages 352--359, Paris, France, June 2019.

\bibitem{Long2015}
Jonathan Long, Evan Shelhamer, and Trevor Darrell.
\newblock {Fully Convolutional Networks for Semantic Segmentation}.
\newblock In {\em Proc. of CVPR}, pages 3431--3440, Boston, MA, USA, June 2015.

\bibitem{Luc2016}
Pauline Luc, Camille Couprie, Soumith Chintala, and Jakob Verbeek.
\newblock {Semantic Segmentation using Adversarial Networks}.
\newblock In {\em NIPS Workshop on Adversarial Training}, pages 1--12,
  Barcelona, Spain, Dec. 2016.

\bibitem{ma2017learning}
Chao Ma, Chih-Yuan Yang, Xiaokang Yang, and Ming-Hsuan Yang.
\newblock {Learning a No-Reference Quality Metric for Single-Image
  Super-Resolution}.
\newblock {\em Computer Vision and Image Understanding}, 158:1--16, May 2017.

\bibitem{Mao2016}
Xudong Mao, Qing Li, Haoran Xie, Raymond Yiu~Keung Lau, Zhen Wang, and
  Stephen~Paul Smolley.
\newblock {Least Squares Generative Adversarial Networks}.
\newblock In {\em Proc. of ICCV}, pages 2794--2802, Venice, Italy, Oct. 2017.

\bibitem{mao2016image}
Xiaojiao Mao, Chunhua Shen, and Yu-Bin Yang.
\newblock {Image Restoration Using Very Deep Convolutional Encoder-Decoder
  Networks With Symmetric Skip Connections}.
\newblock In {\em Proc. of NIPS}, pages 2802--2810, Barcelona, Spain, Dec.
  2016.

\bibitem{minaee2020image}
Shervin Minaee, Yuri Boykov, Fatih Porikli, Antonio Plaza, Nasser Kehtarnavaz,
  and Demetri Terzopoulos.
\newblock {Image Segmentation Using Deep Learning: A Survey}.
\newblock {\em arXiv}, Jan. 2020.
\newblock (arXiv:2001.05566).

\bibitem{Mirza2014}
Mehdi Mirza and Simon Osindero.
\newblock {Conditional Generative Adversarial Nets}.
\newblock {\em arXiv}, Nov. 2014.
\newblock (arXiv:1411.1784).

\bibitem{Noh2015}
Hyeonwoo Noh, Seunghoon Hong, and Bohyung Han.
\newblock {Learning Deconvolution Network for Semantic Segmentation}.
\newblock In {\em Proc. of ICCV}, pages 1520--1528, Las Condes, Chile, Dec.
  2015.

\bibitem{Pan2010}
Sinno~Jialin Pan and Qiang Yang.
\newblock {A Survey on Transfer Learning}.
\newblock {\em IEEE Transactions on Knowledge and Data Engineering},
  22(10):1345--1359, Oct. 2010.

\bibitem{Paszke2016}
Adam Paszke, Abhishek Chaurasia, Sangpil Kim, and Eugenio Culurciello.
\newblock {ENet: A Deep Neural Network Architecture for Real-Time Semantic
  Segmentation}.
\newblock {\em arXiv}, June 2016.
\newblock (arXiv:1606.02147).

\bibitem{Paszke2017}
Adam Paszke, Sam Gross, Soumith Chintala, Gregory Chanan, Edward Yang, Zachary
  DeVito, Zeming Lin, Alban Desmaison, Luca Antiga, and Adam Lerer.
\newblock {Automatic Differentiation in {PyTorch}}.
\newblock In {\em Proc. of NIPS - Workshops}, pages 1--4, Long Beach, CA, USA,
  Dec. 2017.

\bibitem{Romera2018}
Eduardo Romera, Jos\'{e}~M. \'{A}lvarez, Luis~M. Bergasa, and Roberto Arroyo.
\newblock {ERFNet: Efficient Residual Factorized Conv\-Net for Real-Time
  Semantic Segmentation}.
\newblock {\em IEEE Transactions on Intelligent Transportation Systems},
  19(1):263--272, Jan. 2018.

\bibitem{ros2016synthia}
German Ros, Laura Sellart, Joanna Materzynska, David Vazquez, and Antonio~M.
  Lopez.
\newblock {The Synthia Dataset: A Large Collection of Synthetic Images for
  Semantic Segmentation of Urban Scenes}.
\newblock In {\em Proc. of CVPR}, pages 3234--3243, Las Vegas, NV, USA, June
  2016.

\bibitem{Russakovsky2015}
Olga Russakovsky, Jia Deng, Hao Su, Jonathan Krause, Sanjeev Satheesh, Sean Ma,
  Zhiheng Huang, Andrej Karpathy, Aditya Khosla, Michael Bernstein,
  Alexander~C. Berg, and Li Fei-Fei.
\newblock {ImageNet Large Scale Visual Recognition Challenge}.
\newblock {\em International Journal of Computer Vision (IJCV)},
  115(3):211--252, Dec. 2015.

\bibitem{Salomon2004}
David Salomon.
\newblock {\em {Data Compression: The Complete Reference}}.
\newblock Springer Science \& Business Media, 2004.

\bibitem{talebi2018nima}
Hossein Talebi and Peyman Milanfar.
\newblock {NIMA: Neural Image Assessment}.
\newblock {\em IEEE Trans. on Image Processing}, 27(8):3998--4011, Sept. 2018.

\bibitem{Theis2017}
Lucas Theis, Wenzhe Shi, Andrew Cunningham, and Ferenc Huszár.
\newblock {Lossy Image Compression With Compressive Autoencoders}.
\newblock In {\em Proc. of ICLR}, pages 1--19, Toulon, France, Apr. 2017.

\bibitem{tsymbal2004problem}
Alexey Tsymbal.
\newblock {The Problem of Concept Drift: Definitions and Related Work}.
\newblock {\em Computer Science Department, Trinity College Dublin}, 106(2):58,
  Apr. 2004.

\bibitem{ulyanov2017improved}
Dmitry Ulyanov, Andrea Vedaldi, and Victor Lempitsky.
\newblock {Improved Texture Networks: Maximizing Quality and Diversity in
  Feed-Forward Stylization and Texture Synthesis}.
\newblock In {\em Proc. of CVPR}, pages 6924--6932, Honulu, HI, USA, July 2017.

\bibitem{Venkateswara2017}
Hemanth~Demakethepalli Venkateswara, Shayok Chakraborty, and Sethuraman
  Panchanathan.
\newblock {Deep-Learning Systems for Domain Adaptation in Computer Vision:
  Learning Transferable Feature Representations}.
\newblock {\em IEEE Signal Processing Magazine}, 34(6):117--129, Nov. 2017.

\bibitem{visin2016reseg}
Francesco Visin, Marco Ciccone, Adriana Romero, Kyle Kastner, Kyunghyun Cho,
  Yoshua Bengio, Matteo Matteucci, and Aaron Courville.
\newblock {Reseg: A Recurrent Neural Network-Based Model for Semantic
  Segmentation}.
\newblock In {\em Proc. of CVPR - Workshops}, pages 41--48, Las Vegas, NV, USA,
  June 2016.

\bibitem{wang2018high}
Ting-Chun Wang, Ming-Yu Liu, Jun-Yan Zhu, Andrew Tao, Jan Kautz, and Bryan
  Catanzaro.
\newblock {High-Resolution Image Synthesis and Semantic Manipulation With
  Conditional GANs}.
\newblock In {\em Proc. of CVPR}, pages 8798--8807, Salt Lake City, UT, USA,
  June 2018.

\bibitem{wang2004image}
Zhou Wang, Alan~Conrad Bovik, Hamid~Rahim Sheikh, and Eero~P. Simoncelli.
\newblock {Image Quality Assessment: From Error Visibility to Structural
  Similarity}.
\newblock {\em IEEE Trans. on Image Processing}, 13(4):600--612, Apr. 2004.

\bibitem{wang2003multi}
Zhou Wang, Eero Simoncelli, and Alan Bovik.
\newblock {Multi-Scale Structural Similarity for Image Quality Assessment}.
\newblock In {\em Proc. of ACSSC}, pages 1398--1402, Pacific Grove, CA, USA,
  Nov. 2003.

\bibitem{widmer1996learning}
Gerhard Widmer and Miroslav Kubat.
\newblock {Learning in the Presence of Concept Drift and Hidden Contexts}.
\newblock {\em Machine Learning}, 23(1):69--101, Apr. 1996.

\bibitem{yeh2017semantic}
Raymond~A. Yeh, Chen Chen, Teck Yian~Lim, Alexander~G. Schwing, Mark
  Hasegawa-Johnson, and Minh~N. Do.
\newblock {Semantic Image Inpainting With Deep Generative Models}.
\newblock In {\em Proc. of CVPR}, pages 5485--5493, Honolulu, HI, USA, July
  2017.

\bibitem{Rubner2000}
{Yossi Rubner and Carlo Tomasi and Leonidas J. Guibas}.
\newblock {The Earth Mover’s Distance as a Metric for Image Retrieval}.
\newblock {\em International Journal of Computer Vision}, 40(2):99--121, Nov.
  2000.

\bibitem{Yu2018b}
Fisher Yu, Wenqi Xian, Yingying Chen, Fangchen Liu, Mike Liao, Vashisht
  Madhavan, and Trevor Darrell.
\newblock {BDD100K: A Diverse Driving Video Database With Scalable Annotation
  Tooling}.
\newblock {\em arXiv}, Aug. 2018.
\newblock (arXiv:1805.04687).

\bibitem{yu2018generative}
Jiahui Yu, Zhe Lin, Jimei Yang, Xiaohui Shen, Xin Lu, and Thomas~S. Huang.
\newblock {Generative Image Inpainting With Contextual Attention}.
\newblock In {\em Proc. of CVPR}, pages 5505--5514, Salt Lake City, UT, USA,
  June 2018.

\bibitem{zeiler2014visualizing}
Matthew~D. Zeiler and Rob Fergus.
\newblock {Visualizing and Understanding Convolutional Networks}.
\newblock In {\em Proc. of ECCV}, pages 818--833, Zurich, Switzerland, Sept.
  2014.

\bibitem{zeiler2010deconvolutional}
Matthew~D. Zeiler, Dilip Krishnan, Graham~W. Taylor, and Rob Fergus.
\newblock {Deconvolutional Networks}.
\newblock In {\em Proc. of CVPR}, pages 2528--2535, San Francisco, CA, USA,
  June 2010.

\end{thebibliography}
}

\end{document}